\documentclass[runningheads]{llncs}
\usepackage{pdfpages}
\usepackage{graphicx}
\graphicspath{ {images/} }
\usepackage{dsfont}
\usepackage{bm}
\usepackage[utf8]{inputenc} 
\usepackage[T1]{fontenc}    
\usepackage{hyperref}       
\usepackage{url}            
\usepackage{booktabs}       
\usepackage{amsfonts}       
\usepackage{nicefrac}       
\usepackage{microtype}      
\usepackage{amssymb}
\usepackage{amsmath}
\usepackage{multirow}
\usepackage[]{algorithm2e}
\usepackage{afterpage}
\usepackage{float}
\usepackage{listings}

\begin{document}

\title{Uncertainty Weighted Causal Graphs}

%

\author{Eduardo C. Garrido-Merch\'an\inst{1} \and
C. Puente\inst{2} \and
A. Sobrino\inst{3} \and
J.A. Olivas\inst{4}}

\institute{Universidad Aut\'onoma de Madrid, Francisco Tom\'as y Valiente 11, Madrid, Spain
\email{eduardo.garrido@uam.es} \and
Escuela T\'ecnica Superior de Ingenier\'ia ICAI, Universidad Pontificia Comillas, Madrid, Spain
\email{cristina.puente@icai.comillas.edu} \and
Universidad de Santiago de Compostela, A Coru\~na, Spain
\email{alejandro.sobrino@usc.es} \and
Universidad de Castilla La Mancha, Ciudad Real, Spain
\email{joseangel.olivas@uclm.es}
}

\maketitle

\begin{abstract}%
Causality has traditionally been a scientific way to generate knowledge by relating causes to effects. From an imaginery point of view, causal graphs are a helpful tool for representing and infering new causal information. In previous works, we have generated automatically causal graphs associated
to a given concept by analyzing sets of documents and extracting and representing the found causal information in that visual way. The retrieved information shows that causality is frequently imperfect rather than exact, feature gathered by the graph. In this work we will attempt to go a step further modelling the uncertainty in the graph through probabilistic improving the management of the imprecision in the quoted graph.
\end{abstract}

\section{Introduction}
Causality is a key notion in science and philosophy. Physics laws are often expressed in terms of a causal relation, helping in the relevant job of explanation and prediction. For example, Newton’s second law predicts the force necessary (cause) to perform a desired acceleration (effect). In Philosophy, the relevance of causality was highlighted by Aristotle. In Posterior Analytics, he asserted that: we think we have knowledge of a thing only when we have grasped its cause (APost. 71 b 9-11. Cf. APost. 94 a 20). Aristotle also advanced the distinction of causality in four mayor types or classes: material cause, formal cause, efficient cause and final cause.

In the history of thought, Mill \cite{mill1869system} raises the importance of causality in experimental sciences. Inductive inferences from a limited number of observed cases to general instances are feasible because nature is governed by laws. So, the law of universal causation is the guarantee that there are principles that are meant to be discovered if we pursue them actively enough. From a theoretical point of view, our work shares Mill's ideas about the relevance of causality as a condition in the generation of objective scientific knowledge. But as an empirical approach to the analysis of causation, we also adopt Mach’s point of view of causality \cite{mach1976knowledge} as a way of describing, instead of explaining, phenomena: causality is a way to relate cause and effect, not to explain the effect from the cause.

Usually, causality is characterized as a relationship attending the schema \cite{agueda2011causality}: ‘A causes B’, where A is the cause, B the effect and ‘cause’, the causal particle. Traditionally, any causal relationship follows these guidelines \cite{pearl2009causality}:

\begin{itemize}
\item Temporality: causes generally precede their effects.
\item Contiguity: causes are contiguous to the immediate effects.
\item Evidential: causes and effects provide evidence for each other.
To these traditional ideas, we would like to add another:
\item Imperfection: causes, effects and the cause-effect link are
usually qualified by different degrees of strength.
\end{itemize}

This last property is endorsed by the presence of vague words in the aforementioned undisputed properties of causation, as ‘generally precede’, ‘immediate effects’ or ‘is evidence of’. It is a fact that, in many cases, causality is imperfect in nature and causal relations are a matter of degree as we demostrate and try to weight in this work.

In classical logic, deduction is a crisp relation: a conclusion is reached or not from the premises. But as we have previously seen, the explanans sometimes include imprecise generalizations instead of precise laws \cite{apt1990logic}. Therefore, the conclusion or explanadum should be a matter of degree. Imprecise generalizations are common in social sciences and often express tendencies or correlations rather than covering law knowledge.

So according to this imprecise idea of causality, we have organized this paper as follows. First, we introduce theory about causality and related work in this field to provide the context where this work operates to the reader in section 2. Then, we explain how we generate the weighted graph in Section 3, both from the theoretical point of view in the first subsection and technically, in the second subsection of Section 3. Then, we propose a set of synthetic experiments and a real experiment to show the utility of our approach in Section 4. We conclude the paper with a set of conclusions and suggestions for further work in Section 5.

\section{Mining causal sentences to generate knowledge}

In \cite{puente2011text}, Puente, Sobrino, Olivas \& Merlo described a procedure to
automatically display a causal graph from medical knowledge
included in several medical texts.
A morphological and syntactical program was designed to analyze
causal phrases denoted by words like ‘cause’, ‘effect’ or their
synonyms, highlighting vague words that qualify the causal nodes
or the links between them. Another C program received as input a
set of tags from the previous parser generating a template with a
starting node (cause), a causal relation (denoted by lexical words),
possibly qualified, and a final node (effect), possibly modified by a
linguistic hedge showing its intensity. 

\begin{figure}[htb]
\begin{center}
\includegraphics[width=1\linewidth]{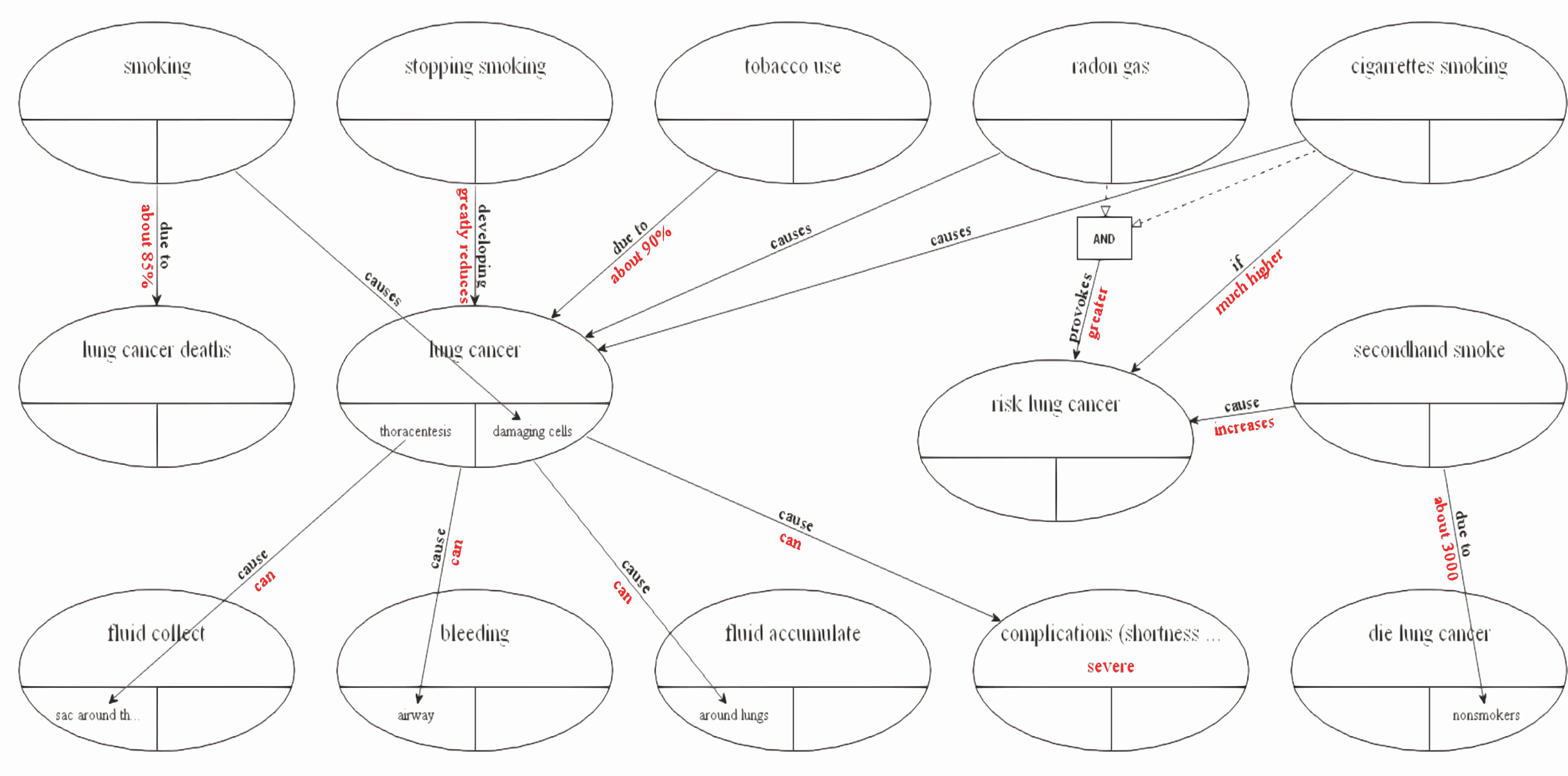}
\end{center}
\vspace{-.7cm}
\caption{Causal representation related to the question What provokes lung cancer?}
\label{fig:syn}
\end{figure}

Once the system was developed, an experiment was performed to
answer the question What provokes lung cancer?, obtaining a set of
causal sentences related to this topic. The whole process was unable to
answer the question directly, but was capable of generating a
causal graph with the topics involved in the proposed question as
shown in Figure \ref{fig:syn}.
The problem with this causal graph is that the weight of each edge is a point estimation of the uncertainty \cite{merchan2019generating}, poorly representing it. We might desire to have a probability distribution in each graph to model its uncertainty. This is so because we analyze several texts
that contain a different certainty degree, represented by an adverb, between the same cause and effect and if we only take point estimations
we lose properties about the uncertainty as symmetry, skewness or kurtosis. That is why we propose to go a step further to evaluate to
what extent a cause provokes and effect, and if so, quantify it.
\vspace{-.3cm}
\section{Generating a weighted graph from Text Causal Relations}
\vspace{-.3cm}
In this section we will introduce theoretically the study about imperfection in causality by means of the extracted causality sentences. Throughout this section, we try to combine two views of the probability. The first one is the probability viewed by logicists, that assigns a certainty factor to each of the rules connecting causes and effects. This view of probability is the one captured by the graph. The other view is the subjective view of probability, followed by the Bayesian community. This view of probability assigns prior distributions and makes Bayesian inference to represent uncertainty. This is the probability that models the weights, or uncertainties, about the certainty factors of the graph. This is the main theoretical difference with respect to bayesian networks \cite{koller2009probabilistic}, that only see probability from a subjective point of view.

\subsection{Introducing uncertainty in certain factors by probability distributions}

In order to model the uncertainty in a more general and principled way \cite{bernardo2009bayesian}, we are going to model each adverb by an univariate probability density function (PDF) over the universe $[0,1]$. Certain factors $\mathbf{x}$ in the graph are the latent variables that we are going to learn about by modelling them with probability distributions whose parametric form, belonging to the exponential family \cite{holland1981exponential}, is given by the retrieved adverbs from the text. This means that, for example, the adverb sometimes is not going to be hard coded as an event with a $50$ percent of probability of happening but with a Gaussian PDF. The process is reversible taking the MAP of the distribution: 
\vspace{-.1cm}
\begin{align}
x_{MAP} = \arg\max_{x \in \Omega} P(x).
\end{align}
Where $\Omega$ is the space $[0,1] \in \mathcal{R}$. It is interesting to observe that by representing the adverbs with probability distributions we are proposing a generalization of the point estimation model. We are going to model each adverb with Gaussian, Beta and Exponential distributions.  For example, consider the adverb \textit{hardly ever} w.r.t \textit{sometimes}. When we qualify a causal relation by using the adverb \textit{sometimes} the uncertainty is broad. It can occur with a $25\%$ or a $75\%$ of probability. On the other hand, when using \textit{hardly ever} the uncertainty interval is low, maybe just a $10\%$ or a $20\%$ of probability. Just by representing \textit{hardly ever} by a point estimate we cannot represent this information. Probability distributions fit this scenario, assigning a probability mass to every possible certainty factor in the $[0,1]$ universe.

\begin{figure}[htb]
\begin{center}
\begin{tabular}{cc}
\includegraphics[width=0.49\linewidth]{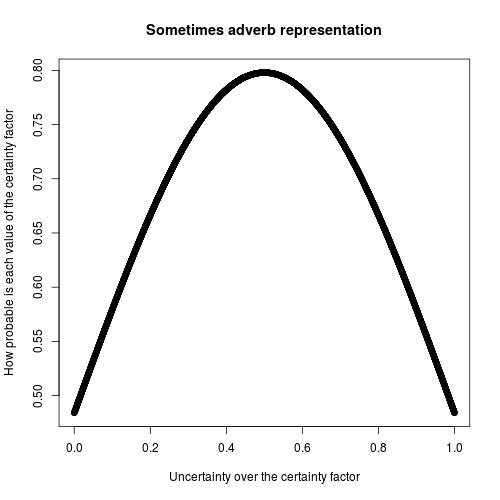} &
\includegraphics[width=0.49\linewidth]{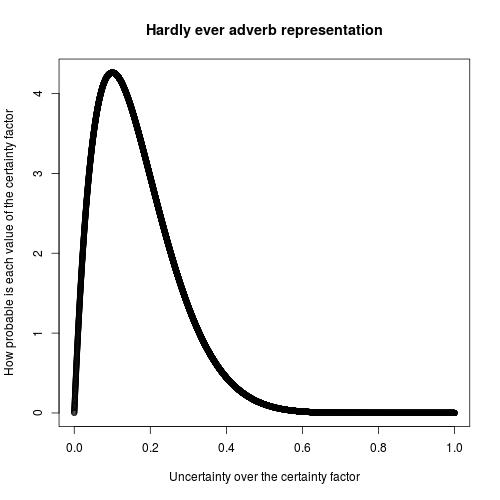}\\
\end{tabular}
\caption{Prior distribution of two of the considered time adverbs.}
\label{fig:adverbs}
\end{center}
\end{figure}

It is common that causal relations come qualified with adverbs such as \textit{Always} and we could think in placing a Delta Distribution over $1$ but as David Hume said about causality \cite{hume2011letters} we can not be sure about the future effects of a cause just by the past effects of this cause, the induction principle is arguable, so we place a spiky exponential distribution just in case the causal relation does not hold in the future by the random law of probability of rare events. We can observe these distributions in Figure \ref{fig:adverbs}.
Causal relations can appear in a multiple number of texts relating the same relation $x$ but qualified by a different time adverb $a_i,a_j \in A$, where $A$ is the set of all possible adverbs. In this case, we are interested in inferring a posterior distribution $p(x|a_i,a_j)$ over the certain factor latent variable of the relation $x$ that represents a mixture of the uncertainties of the analyzed time adverbs $p(x|a_i), p(x|a_j)$. We propose a learning method to compute such posterior distribution $p(x|a_1,...,a_n)$ for every certain factor latent variable.

\begin{figure}[htb]
\begin{tabular}{ccc}
\includegraphics[width=0.49\linewidth]{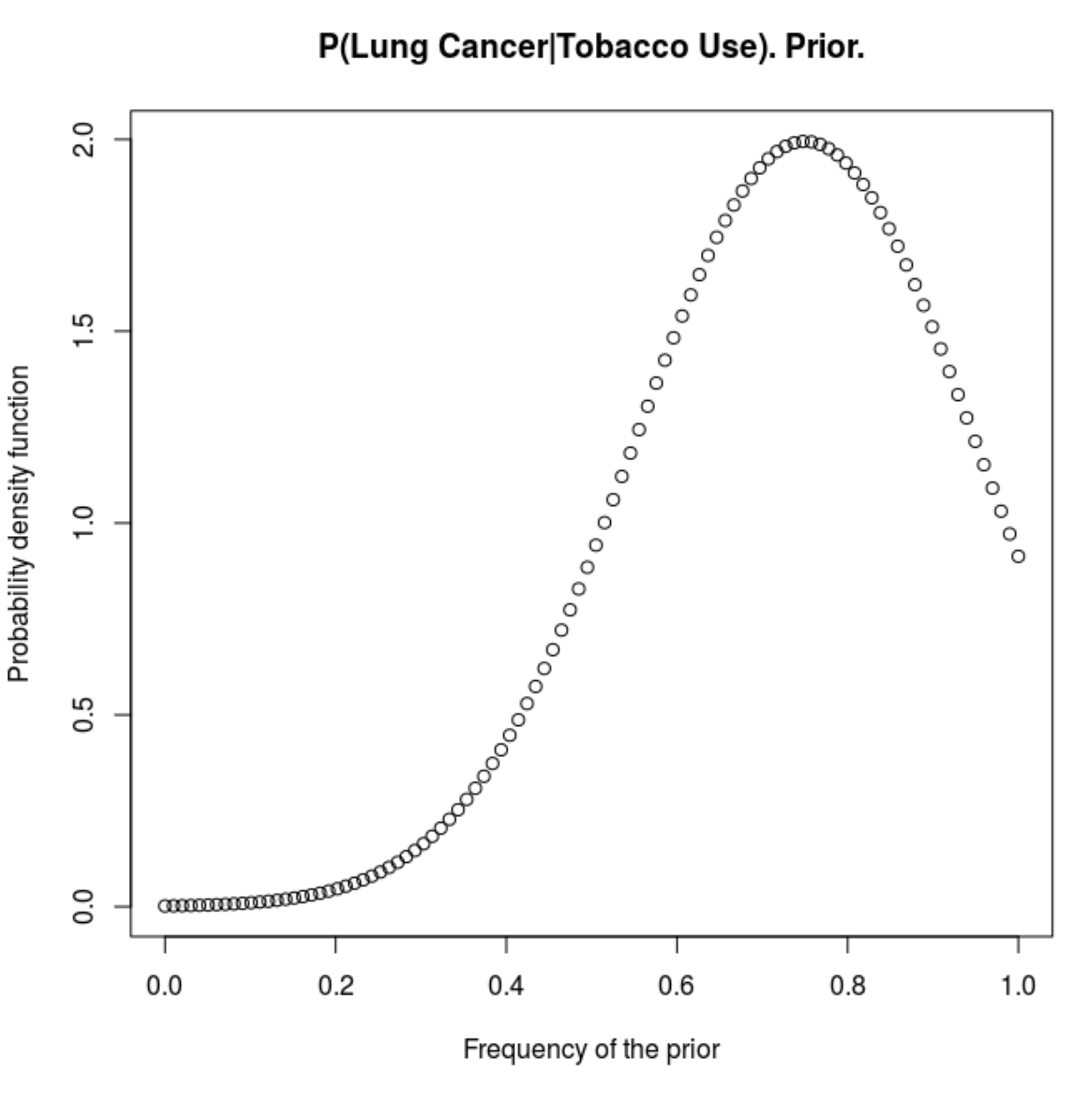}&
\includegraphics[width=0.49\linewidth]{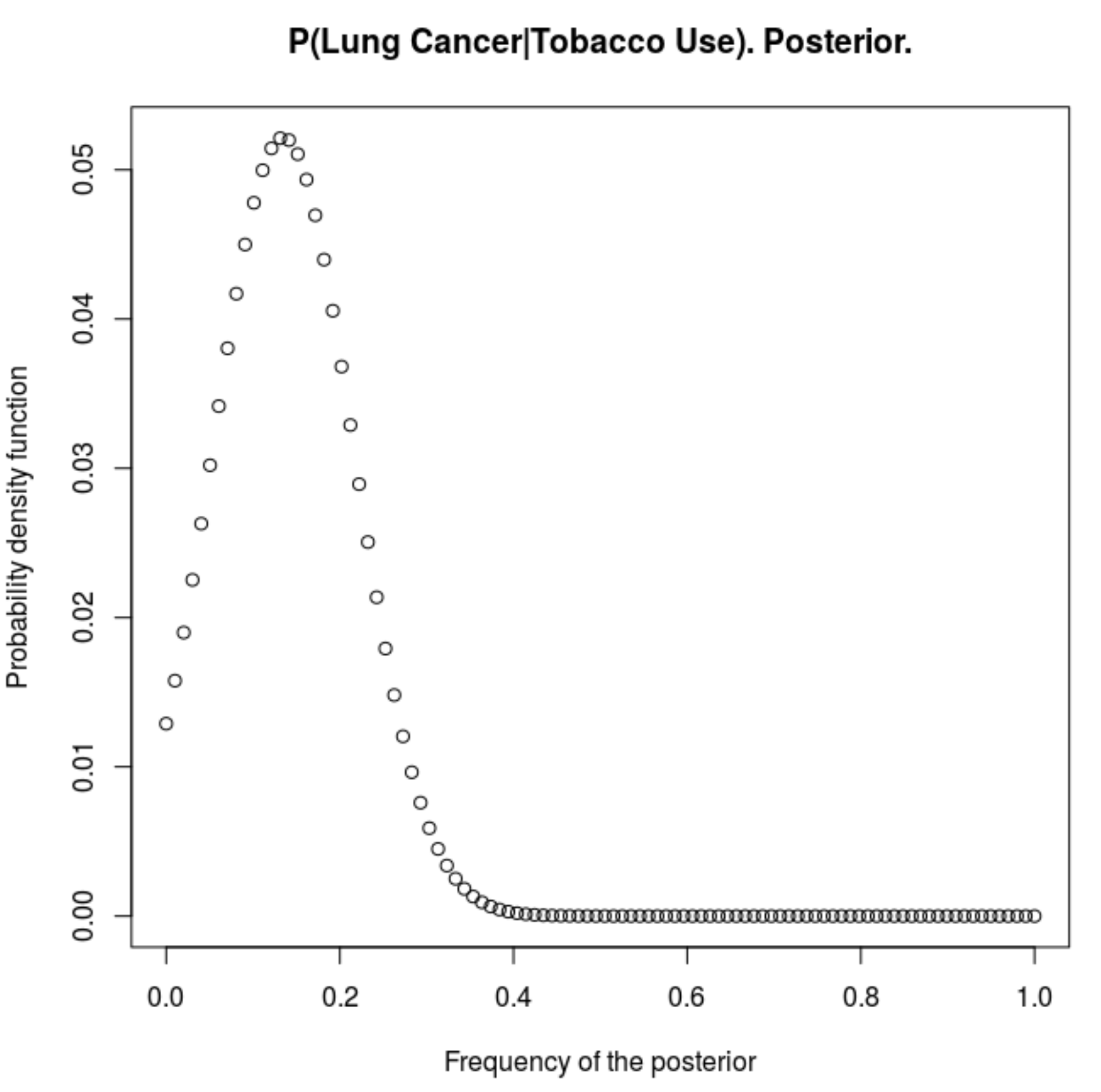}\\
\end{tabular}
\caption{Learning process. (Left) Prior distribution. (Right) Posterior distribution.}
\label{fig:learning}
\end{figure}

\begin{figure}[htb]
\begin{center}
\begin{tabular}{cc}
\includegraphics[width=0.45\linewidth]{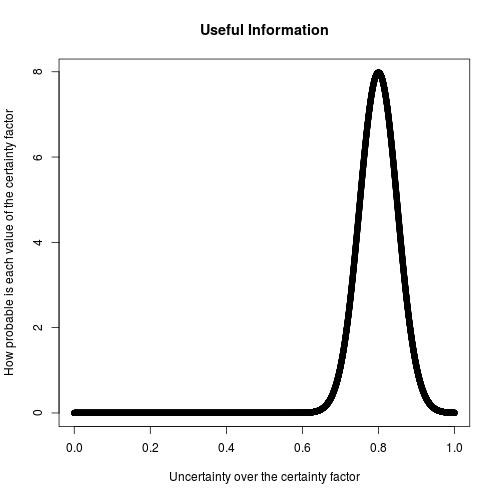}&
\includegraphics[width=0.45\linewidth]{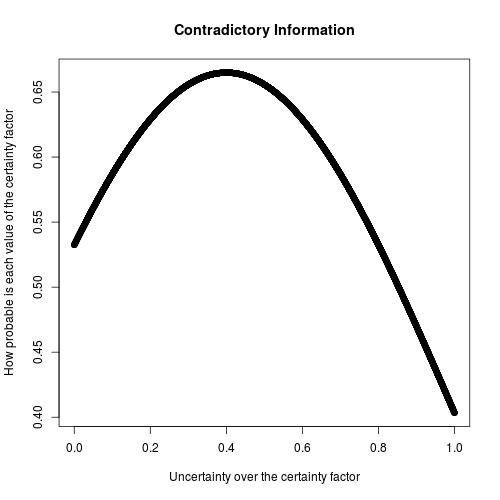}\\
\end{tabular}
\caption{Posterior distributions about the certainty factors latent variables. (Left) Low entropy, useful information. (Right) High entropy, contradictory information.}
\label{fig:posteriors}
\end{center}
\end{figure}

First, we infer the posterior distribution $p(x)$ between two prior distributions by multiplication $p(x|a_i)p(x|a_j)$. In order to do so, we discretize the value of this distributions in the $[0,1]$ universe by approximating them using a grid $\mathbf{g} \approx \mathcal{X} = \mathcal{R}^{[0,1]}$, which is commonly defined as a Grid Approximation. The Grid Approximation method for multiplying distributions is inadvisable for Multivariate Distributions \cite{mcelreath2018statistical} but as we are considering Univariate Distributions is a plausible method. After the multiplication we normalize the posterior distribution by dividing each value in the grid by the sum of all the values in the grid $p(x_k) = p(x_{k-1}) / \sum_{k=1}^N p(x_k)$. We have considered this process since the resultant distribution will have lower entropy than the previous two distributions if these distributions are similar. If we analyze a high number of causal relations and all the time adverbs are similar then we are interested in concluding that we are sure about the uncertainty degree over the certainty factor latent variable \ref{fig:learning}.
Moreover, if the time adverbs are not similar, we have learned contradictory information about the certainty factors \ref{fig:posteriors}.
If we want to generate text that represent the learned PDF we need to find the adverb that best represent the posterior distribution $p(x|a_1,...,a_n)$. We do so by minimizing the KL Divergence between the posterior and the adverbs. The KL divergence is a measure of distance between distributions \cite{murphy2012machine}. 
\vspace{-.3cm}
\begin{align}
a^\star = \arg\min_{a \in \mathbf{A}} KL(a || P).
\end{align}
Where $a$ represent the PDF of an adverb, $P$ the learned posterior and $\mathbf{A}$ is the space of all the considered adverb priors. The KL divergence between two distributions $P$ and $Q$ is given by the following expression, where we approximate the integral by Grid Approximation:
\begin{align}
D_{KL}(P || Q) = \int_{\infty}^{\infty} p(x)\log(\frac{p(x)}{q(x)}) dx.
\end{align}
In order to infer conditional distributions in the graph we just multiply the PDFs that relate the intermediate concepts. We can find the adverb that best represents the joint probability distribution by minimizing the KL Divergence. For example, let us consider that a cause $a$ produces and effect $b$ and the effect $b$ produces another effect $c$. We have learned $P(b|a)$ and $P(c|b)$ by multiplying distributions. So we now need to compute $P(c|b,a)$, this expression is just given by $P(c|b,a) = P(c|b)P(b|a)$. This expression generalizes for any potential chain of events in the graph. We can also take a point estimation for $P(z|x_1,...,x_n)$ by computing its MAP. If we think that $P(z|x_1,...,x_n)$ may be multi-modal we can resort to a sampling algorithm as the Metropolis Hastings algorithm and then compute a sufficient statistic as the mean or median.

It is important to denote that this pondered graph is not a Bayesian Network. In Bayesian Networks the probability distributions are associated with the nodes of the graph \cite{jensen1996introduction} and in this case the probability distributions are associated with the edges of the graph. For the sake of understanding we have denoted the probability distribution between the nodes $a$ and $b$ as $P(b|a)$ but this is not the relation that is established in Bayesian Networks. Here, we do not know the probability of $a$ as the node $a$ does not represent a random variable but a cause or effect in the sense that logicists view an effect caused by a rule. What we are modelling is the uncertainty in the causal relation between concepts $a$ and $b$. This uncertainty is the one that is represented by $P(b|a)$ and represents the uncertainty over each possible certainty factor between the concepts $a$ and $b$.

The applications of our proposed model and bayesian networks are different. This model can be used, as we will further see in the experiments section, to represent the uncertainty over learned sets of rules each of which connects the cause $a$ and effect $b$ by a different certainty factor $c \in [0,1]$. As far as we know, no model is able to provide as estimation of $p(b|a)$ in the logicist point of view of probability for the certainty factor $c$ that connects $b$ and $a$, only a maximum a posteriori estimation $c$, that is equal to the maximum of the probability distribution that our model infers and proposes as a novel contribution: $c = \arg \max p(b|a)$. This uncertainty between $a$ and $b$ of the certain factors, among which we provide a probability distribution, is the result of being modelled by a random variable $\epsilon$ that defines a probabilistic space $(E,\Omega,P)$, being $E$ the set of all events and $\Omega$ the $\sigma$-algebra of $E$. We model each different adverb found for the same cause and result by a different probability distribution and make inference of them all but the surrogate random variable between cause and effect is the same.

This does not happen in bayesian networks, where each node represents a random variable. Bayesian Networks are probabilistic graphical models that model conditional independence, they also model causation but in a different way, in these models, the edges represent conditional dependence and they are not a random variable. The nodes of the graph are the random variables that define a factorized representation of the joint probability distributiontaking into account the conditional independences defined by the graph. If and edge between $a$ and $b$ exist, it means that $p(b|a)$ is a factor in the joint probability distribution. Then, knowing values of $a$ and $b$, we can conduct inference. We can see that the applications of bayesian networks are totally different from the ones of our proposed model. While bayesian networks perform inference with respect to different random variables in a graphical model to represent the global uncertainty of the joint probability distribution of the model, our method perform inference of the random variable modelling the uncertainty of each certain factor to perfectly represent the uncertainty of each causal relation. Bayesian networks are pure subjetivist models and our models are a mixture between logicists and subjetivist models.

We now generate the same inference procedure for every pair of retrieved connected nodes by an edge. This generates a pondered graph that connects every pair of nodes by a probability distribution. We can now compute the probability distribution of nodes connected by two or more edges by the product of probability distributions of all the edges that connect them. Suppose that two nodes $a$ and $c$ are connected by a set of nodes $\mathbf{b}$ , which we can denote $P(c|a,b_1,...,b_n)$, then, this is equivalent to connect $a$ and $c$ with the probability distribution $P(c|a) = \sum_{i=2}^{N} P(b_i|b_{i-1}) P(b_1)  P(a)$, where $N$ are the number of nodes between $a$ and $c$.

\begin{figure}[htb]
\begin{center}
\includegraphics[width=0.7\linewidth]{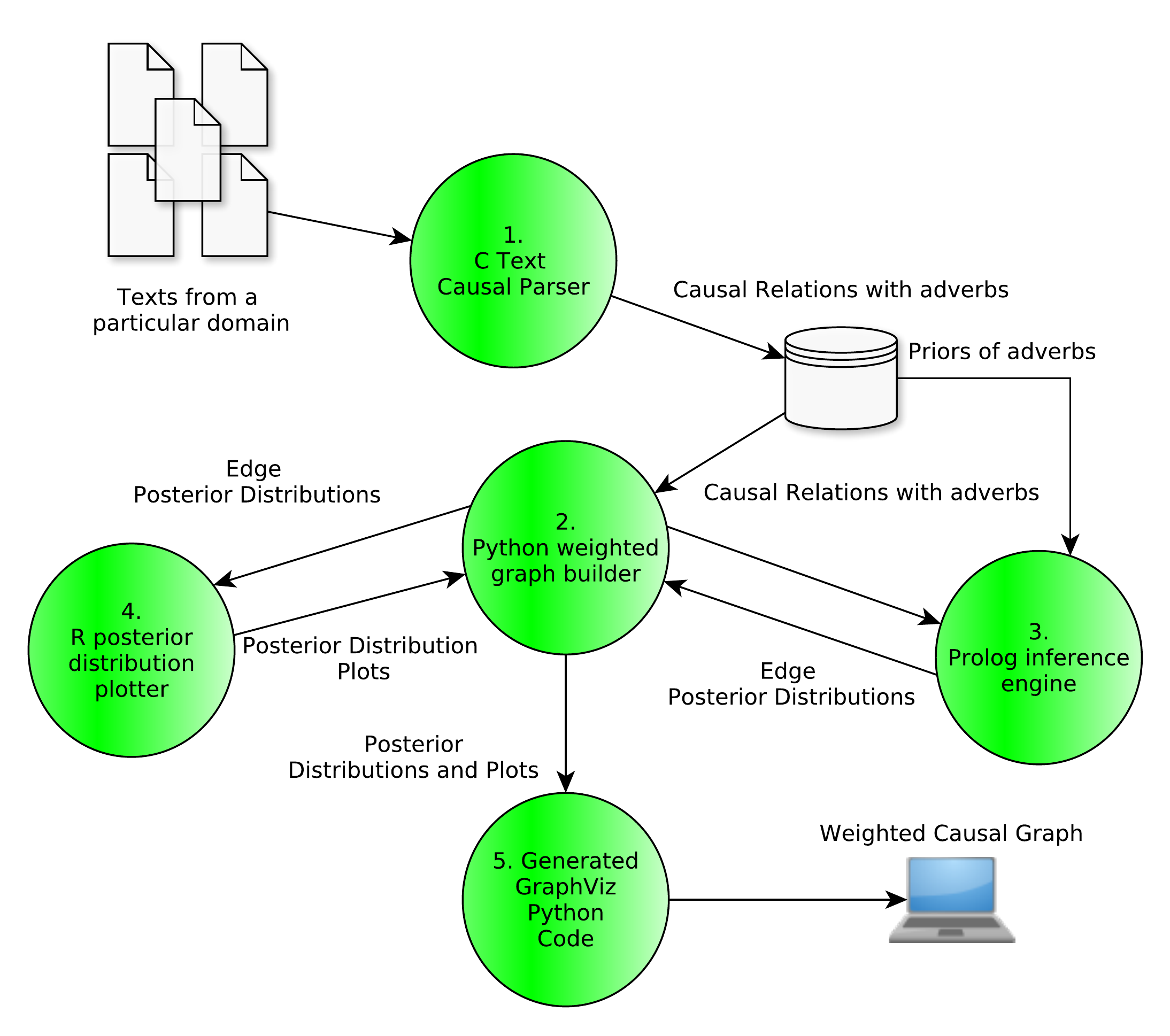}
\end{center}
\vspace{-.7cm}
\caption{Flow diagram of our proposed system architecture.}
\label{fig:dfd}
\end{figure}

Computing these posteriors is useful to answer questions that involve concepts whose causal relations does not explicity relate them, but implicity, they are connected. Figure \ref{fig:dfd} includes the architecture of our system that creates automatically the pondered graphs for visualization and representation of the uncertainty.
We can see that our model represents the uncertainty over the logicist certainty factors. Previous models only output a certain factor and our proposed model give one additional dimension of uncertainty, better modelling it. Applications involve a better decision analysis with respect to causation between nodes, a better representation of the uncertainty when several different retrieved causal relations involving different adverbs between a cause and an effect exist in texts and a visualization of the causation between nodes in a graph and a treatment of fake news or fake information that is retrieved by several sources detected by high entropy posterior distributions or high KL divergence with respect to a new causal relation. 
\section{Experiments}
In this section we will show the usefulness of our proposed approach in a set of synthetic experiments and in a real experiment involving lung cancer. These experiments provide empirical claims to support our hypothesis that a pondered graph can be generated from a set of texts belonging to a particular domain. 
\subsection{Synthetic Experiments}
\begin{figure}[htb]
\begin{center}
\begin{tabular}{cc}
\includegraphics[width=0.475\linewidth]{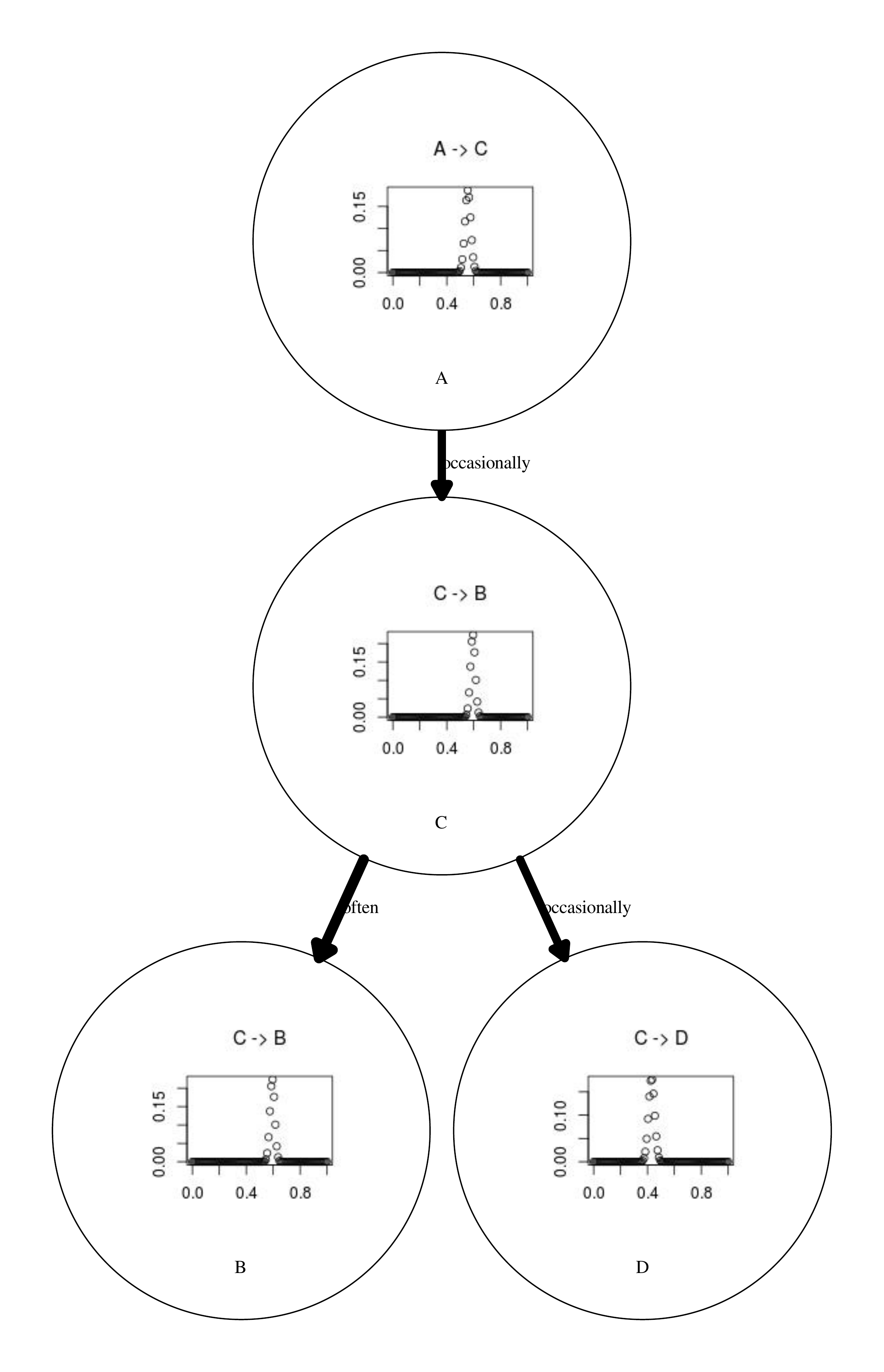}&
\includegraphics[width=0.475\linewidth]{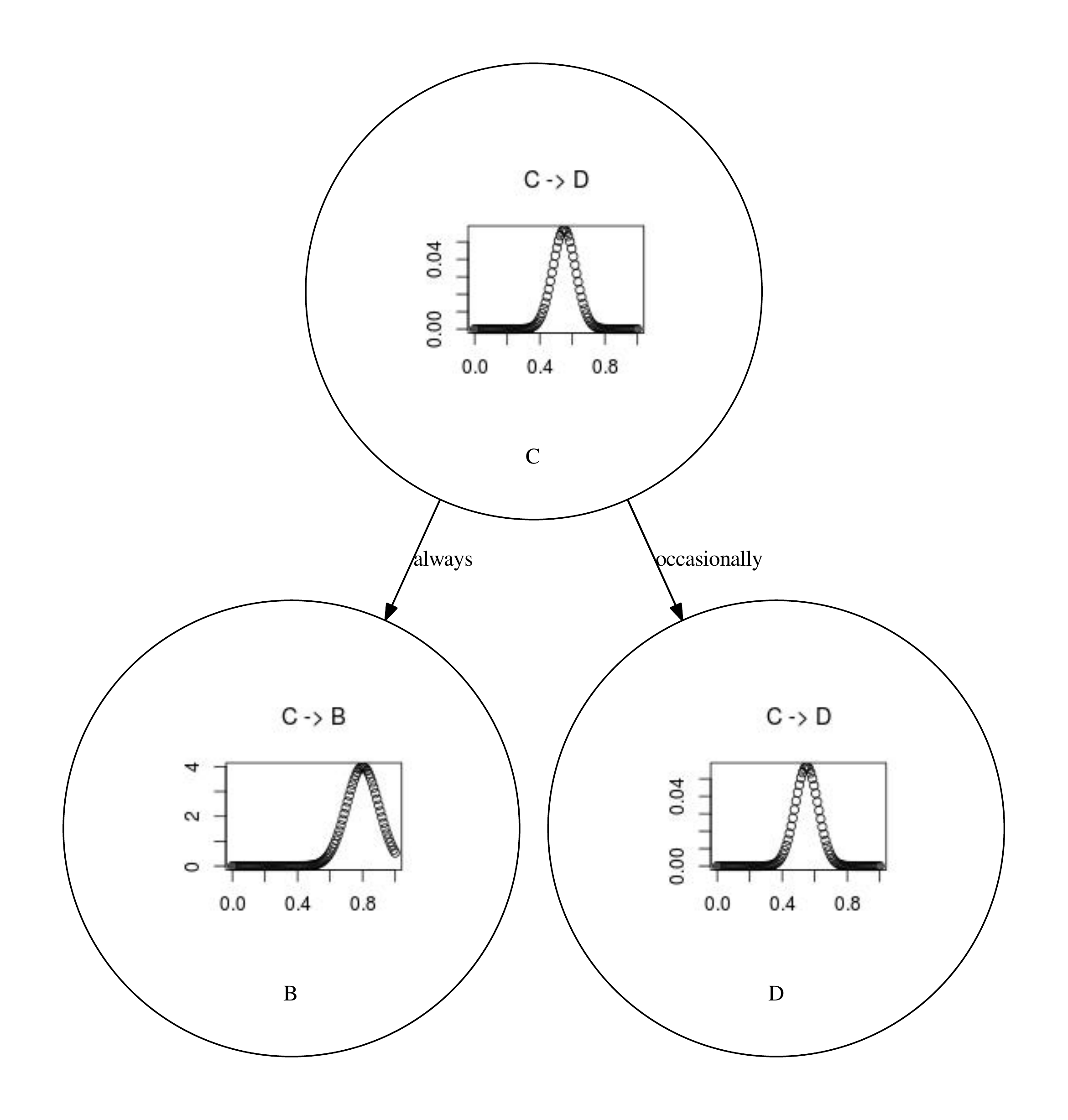}\\
\end{tabular}
\end{center}
\vspace{-.7cm}
\caption{Learned pondered graphs of a toy problem.}
\label{fig:toy}
\end{figure}

It is interesting to see how our tool displays the causal information described in previous sections. Just to show the pondered graphs that the system is capable of generating, we create a toy problem where we are going to sample the following causal relations: \textit{A -> C. C -> D. C -> B} where \textit{A,B,C,D} are the nodes of the causal graph. We configure the generation script to sample random causal relations between all the \textit{A,B,C,D} nodes and retain only the causal relations \textit{A -> C. C -> D. C -> B}. We configure the data generation script to generate different number of causal relations just to see how the posterior distributions behave after being computed from a different number of causal relations. We also configure an scenario where the sampled adverbs are always similar where we expect the posterior distributions to have low entropy. We show the learned pondered graphs of this toy problem in Figure \ref{fig:toy}.
We can see in Figure \ref{fig:toy} different pondered graphs that the system has learned by processing the causal relations that the data generator file has generated. The size of the edge is proportional to the number of analyzed causal relations between the nodes that are connected by the edge. The adverb that is plotted in the edge is the adverb whose probability distribution minimizes the KL divergence with respect to the inferred posterior distribution of that edge.

 In the first pondered graph we can see how the probability distributions are spiky, i.e., have low entropy. These distributions have been inferred by causal relations that contained very similar adverbs, so the prior distributions associated with these adverbs were similar. We can say in this scenario that we are sure about the learned knowledge, or at least, that we are certain that our uncertainty over the acquired knowledge is low. On the other hand, in the second scenario we have generated a very small number of causal relations, so the output did not even acquire all the valid causal relations and the uncertainty over the causality of these concepts is very high. The computational time of this process is quadratic over the number of causal relations and the size of the grid that we use to approximate the posterior distributions, that is: $\mathbb{O}(NR)$ where $N$ is the number of causal relations and $R$ is the size of the grid. 
\vspace{-.5cm}
\subsection{Real Experiment}
\begin{figure}[htb]
\begin{center}
\includegraphics[width=1\linewidth]{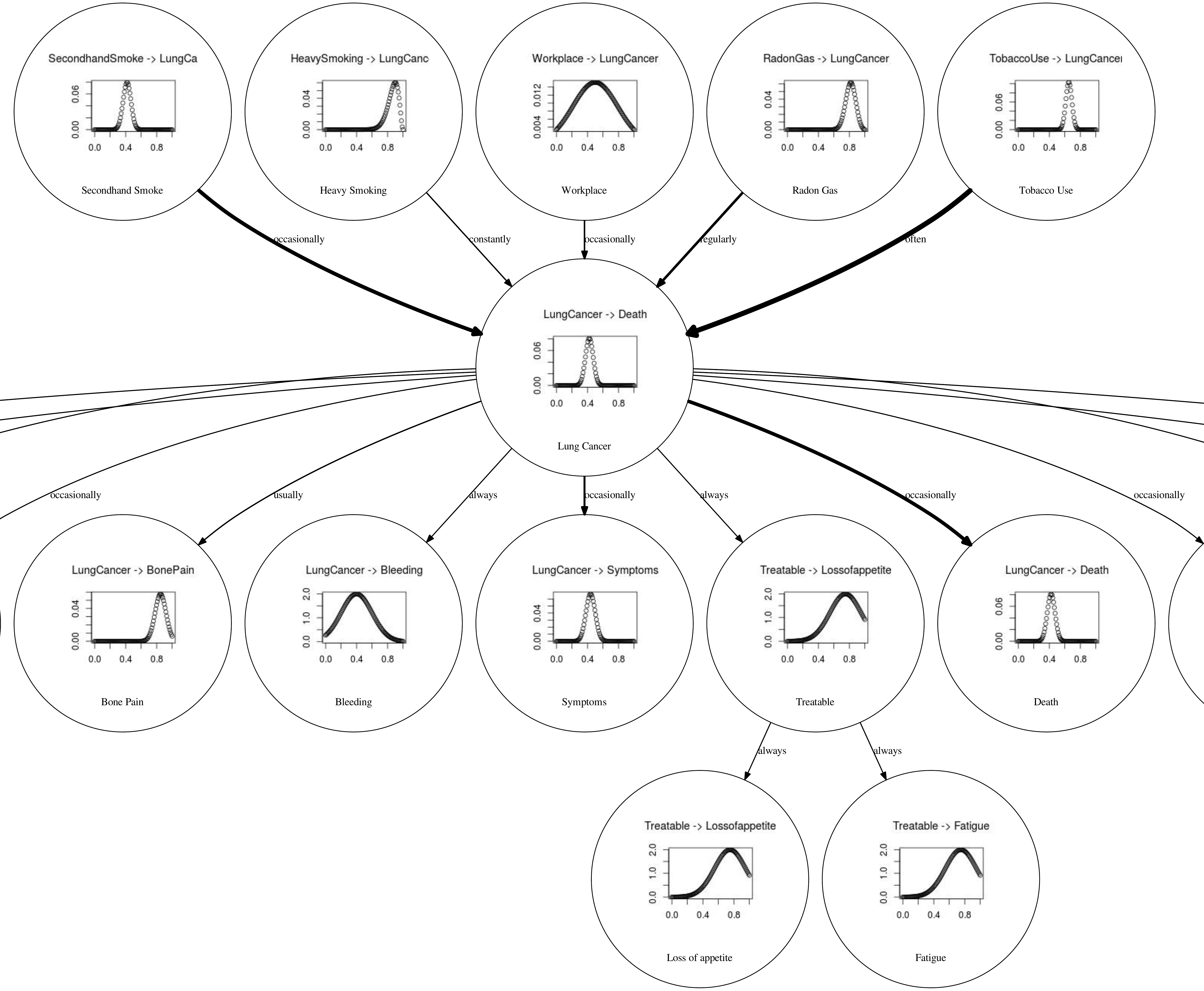}
\end{center}
\vspace{-.7cm}
\caption{Partial section of the causal graph computed from the lung cancer scenario.}
\label{fig:real}
\end{figure}

Having tested the system in the previous section with a set of synthetic experiments we are now interested in this section in showing a real case scenario where our approach can provide an elegant representation of the causal knowledge hidden in a set of texts. We have analyzed several text from the lung cancer domain using the mentioned architecture. By analyzing those texts, our system has inferred the graph whose main part is shown in Figure \ref{fig:real}. We have cut the other causal relations of the graph for the sake of visibility. 
We can observe a great variety of posterior distributions that connect every cause and effect in the graph. For example, we are certain the lung cancer can produce death half of the times or that inhaling radon gas is a frequent cause of developing lung cancer. We have also learn that the probability degree of developing lung cancer with respect to the workplace is very uncertain. We can see the accuracy of our system in that the probability distribution of the causal relation that involve heavy smoking with lung cancer resides at the left (almost close to 1) that the probability distribution of smoking with lung cancer. This validates the hypothesis that the amount of smoking is directly proportional to developing lung cancer and that our system is capable of inferring posterior probability distributions that are coherent with prior premises. 
\section{Conclusions and further work}
We have proposed a methodology that generates pondered causal graphs from sets of analyzed texts of particular domains. By pondering the causal relations that appear in these texts we have represented, with probability distributions, the uncertainty involving these causal relations. We have illustrated the theoretical and practical details of our methodology. Our system serves as a representation of the adquired knowledge and as a question and answering system \cite{puente2013answering} with uncertainty.

In order to also represent vagueness we will resort to fuzzy logic \cite{yen1999fuzzy} and build a similar methodology to compare both approaches. We will also propose a system that uses the model to tackle fake news \cite{lazer2018science}. Summarizing the causal information \cite{puente2013creating} \cite{puente2017summarizing} \cite{puente2015summarizing} with the learned uncertainty is also a pending task. Another interesting line of work is proposing an evaluation measure of the graph and then optimizing the priors and types of the distributions to detect which are the best prior distributions their hyperparameters. In order to do so we will resort to multiobjective Bayesian Optimization with Constraints \cite{garrido2019predictive}. Finally, we plan to integrate the weighted causal graph in a machine consciousness architecture implemented in robots \cite{merchn2020machine} in to make them question and answer questions from its knowledge base, exhibiting human behaviour.

\section*{Acknowledgments}

The authors gratefully acknowledge the use of the facilities of Centro
de Computaci\'on Cient\'ifica (CCC) at Universidad Aut\'onoma de
Madrid. The authors also acknowledge financial support from Spanish
Plan Nacional I+D+i, grants TIN2016-76406-P and TEC2016-81900-REDT.

\bibliographystyle{acm}
\bibliography{bayesian_graph}
\end{document}